\documentclass[letterpaper, 10 pt, conference]{ieeeconf}  

\IEEEoverridecommandlockouts                              

\usepackage[noadjust]{cite}
\usepackage{amsmath}
\usepackage{amssymb}
\usepackage{graphicx}
\usepackage{subcaption}
\usepackage[table,xcdraw,dvipsnames]{xcolor}
\usepackage[bookmarks=true,colorlinks=true]{hyperref}
\newcommand{\minus}{\text{-}}

\title{\LARGE \bf CropNav: a Framework for Autonomous Navigation in Real Farms}

\author{Mateus V. Gasparino$^{1,2}$, Vitor A. H. Higuti$^{2}$, Arun N. Sivakumar$^{1,2}$,\\
    Andres E. B. Velasquez$^{1}$, Marcelo Becker$^{3}$ and Girish Chowdhary$^{1}$
\thanks{*This work was supported in part by NSF STTR Phase 2 \#1951250 and Agriculture and Food Research Initiative (AFRI) grant no. 2020-67021-32799/project accession no.1024178 from the USDA National Institute of Food and Agriculture: NSF/USDA National AI Institute: AIFARMS, Mateus V. Gasparino and Arun N. Sivakumar were interns at EarthSense during May-Aug 2022. }
\thanks{Project website: \url{https://mateusgasparino.com/cropnav}}%
\thanks{$^{1}$Field Robotics Engineering and Science Hub (FRESH), Illinois Autonomous Farm, University of Illinois at Urbana-Champaign (UIUC), IL}%
\thanks{$^{2}$EarthSense Inc., Champaign, IL, USA}%
\thanks{$^{3}$Dept. of Mechanical Engineering, University of São Paulo (USP), São Carlos, SP, Brazil}%
\thanks{{Correspondence to \tt\small \{mvalve2,girishc\}@illinois.edu}}%
}

\begin{document}

\maketitle
\thispagestyle{empty}
\pagestyle{empty}

\begin{abstract}

Small robots that can operate under the plant canopy can enable new possibilities in agriculture. However, unlike larger autonomous tractors, autonomous navigation for such under canopy robots remains an open challenge because Global Navigation Satellite System (GNSS) is unreliable under the plant canopy. We present  
a hybrid navigation system that autonomously switches between different sets of sensing modalities to enable full field navigation, both inside and outside of crop. 
By choosing the appropriate path reference source, the robot can accommodate for loss of GNSS signal quality and leverage row-crop structure to autonomously navigate. However, such switching can be tricky and difficult to execute over scale. Our system provides a solution by automatically switching between an exteroceptive sensing based system, such as Light Detection And Ranging (LiDAR) row-following navigation 
and waypoints path tracking. In addition, we show how our system can detect when the navigate fails and recover automatically extending the autonomous time and mitigating the necessity of human intervention. Our system shows an improvement of about 750~m per intervention over GNSS-based navigation and 500~m over row following navigation.


\end{abstract}


\section{INTRODUCTION}

Robots are increasingly being used in agriculture \cite{english2014vision,bergerman2015robot,emmi2021hybrid,cantelli2019small,higuti2019under,kayacan2018high,mueller2017robotanist,velasquez2021multi}. In large scale commodity row-crops, such as corn and soybean, such small robots can be beneficial for a variety of tasks, such as phenotyping data acquisition, crop scouting for disease and other stressors, under canopy cover crop planting, and under canopy weeds removal \cite{McAllister2020RSS,khanna2022digital,r2018research}. These tasks cannot be performed by bigger equipment, because they are designed for open-field or over the canopy operations. As such, it is increasingly clear that small autonomous mobile robots provide the necessary compact and low-cost form factor appropriate machines for the aforementioned key tasks. However, to enable the full potential for small agbots, autonomous navigation algorithms that can handle challenging field scenario, under-canopy clutter, and tight spaces are required \cite{underwood2017efficient, mueller2017robotanist, kayacan2018high,higuti2019under}. Given the large scale of row-crops agriculture, the algorithms must be capable of performing in a variety of situations, and over larger distances than previously reported in the literature \cite{english2014vision,bergerman2015robot,emmi2021hybrid,cantelli2019small,higuti2019under,kayacan2018high,mueller2017robotanist,velasquez2021multi}.



\begin{figure}[t]
    \centering
    \includegraphics[width=0.9\linewidth]{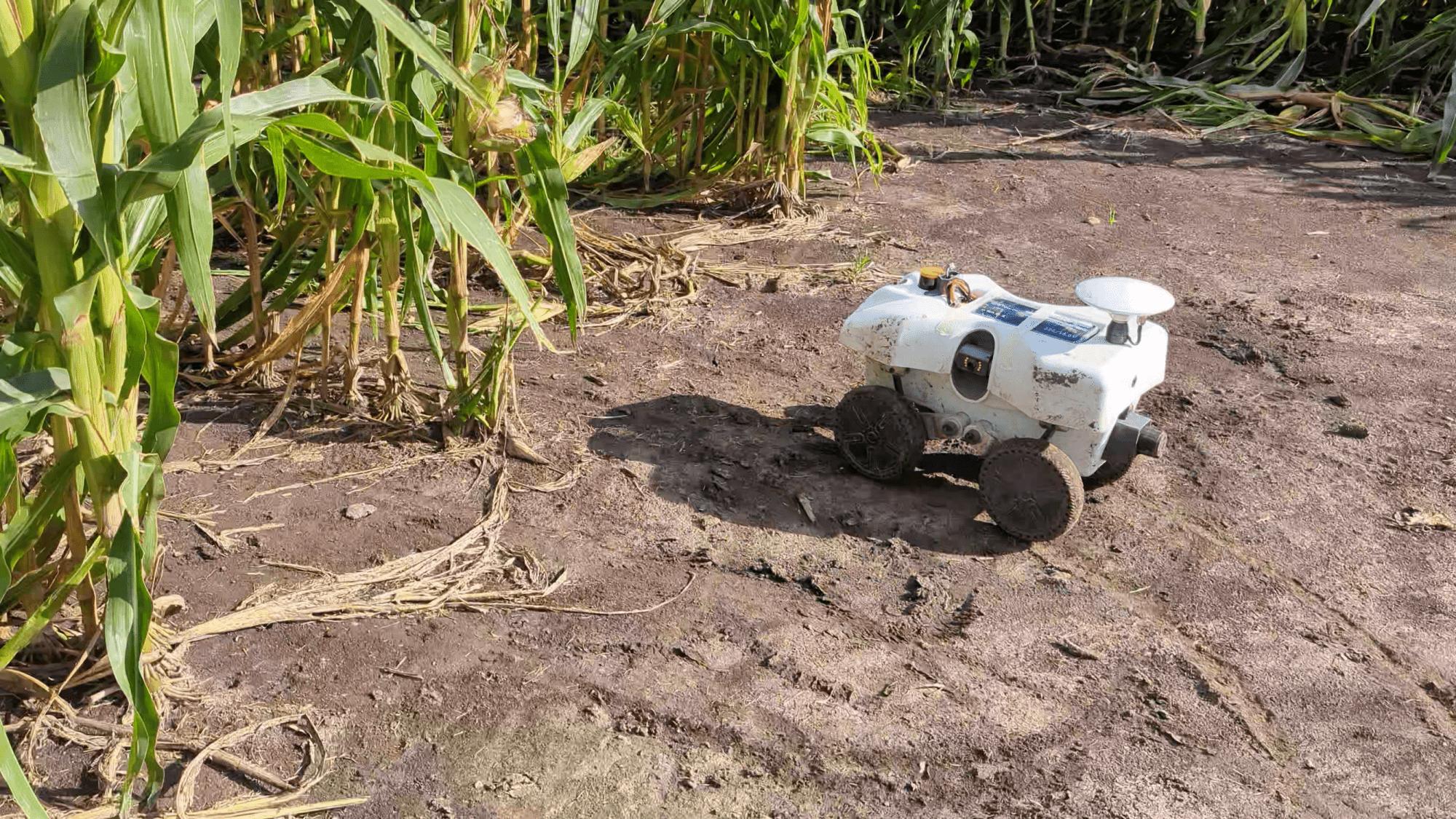}
    \caption{We present a solution to autonomously navigate in farm environments. By fusing multiple sensors, our system is able to smartly switch between reference modalities to provide safe navigation in real agricultural scenarios.}
    \label{fig:terrasentia}
    \vspace{-0.2in}
\end{figure}

Autonomous platforms need to ensure accurate localization and perception to avoid damaging the crop. This necessitates the use of various embedded sensors and sensory fusion techniques to extract environment data and control the vehicle \cite{gasparino2020improved,velasquez2021multi,ball2016vision}. Currently, most works focus on two main types of navigation, GNSS-based path tracking \cite{kayacan2018high,cantelli2019small} and row follow navigation \cite{higuti2019under,velasquez2021multi,gasparino2020improved,sivakumar2021learned}. Due to its wide availability, GNSS is often used in outdoor applications such as agricultural environments. With clear line of sight to the satellites, GNSS can provide an estimate for the vehicle's absolute position. Furthermore, combined with the Real-Time Kinematic hardware (RTK), GNSS can provide position with an accuracy of few centimeters. Indeed, several agricultural robotic systems utilize GNSS and RTK technologies for autonomous navigation \cite{kayacan2018high,rovira2015role,reid2000agricultural,bak2004agricultural,bakker2011autonomous,zhang2017development}. However, GNSS, even with RTK corrections, does not work well in occluded environments, such as under plant canopy, due to multi-path errors. As such, these systems for autonomous navigation in between crop-rows is fraught with failures. To circumvent this issue, many works use exteroceptive sensors to navigate in between crop rows \cite{higuti2019under,velasquez2021multi,gasparino2020improved,sivakumar2021learned}. These methods are designed to leverage features found in regularly planted crop rows. However, such features or other distinguishable landmarks could often be missing when the robot is out of the row and maneuvering in the headlands. Therefore, purely exteroceptive methods cannot cover the entire field and are limited to specific scenarios. 

To address the shortcomings of pure GNSS-based navigation or pure exteroceptive row-following, we present a hybrid autonomy architecture that can leverage the strengths of both. Our key contributions are in architecting this system to ensure reliable long-distance navigation in real agricultural fields through our collision detection based on friction coefficients and switching mechanism between under-canopy and open space navigation. Unlike previous methods \cite{kayacan2018high, gasparino2022wayfast, velasquez2021multi,sivakumar2021learned,higuti2019under}, we present a framework that exploit GNSS-based navigation when convenient. Our approach is modular, utilizing powerful state-of-the-art techniques in field robot navigation, including predictive controllers, receding horizon estimators, and sensor fusion techniques \cite{kayacan2018high, gasparino2022wayfast, velasquez2021multi, mayne2014model,alessandri2003receding}. It is capable of utilizing feature based row-followers such as \cite{higuti2019under,velasquez2021multi,gasparino2020improved} or learning based row-followers such as \cite{sivakumar2021learned,zou2019robust,neven2018towards,chang2018deepvp,zhang2012monocular}. When a collision is identified, our method is able to recover for failures and improve navigability in challenging agricultural fields. The approach is evaluated on long paths in real agricultural fields, and is shown to significantly outperform GNSS only navigation.

\begin{figure*}[t]
    \centering
    \includegraphics[trim={0cm 5cm 0 5cm}, clip, width=0.8\linewidth]{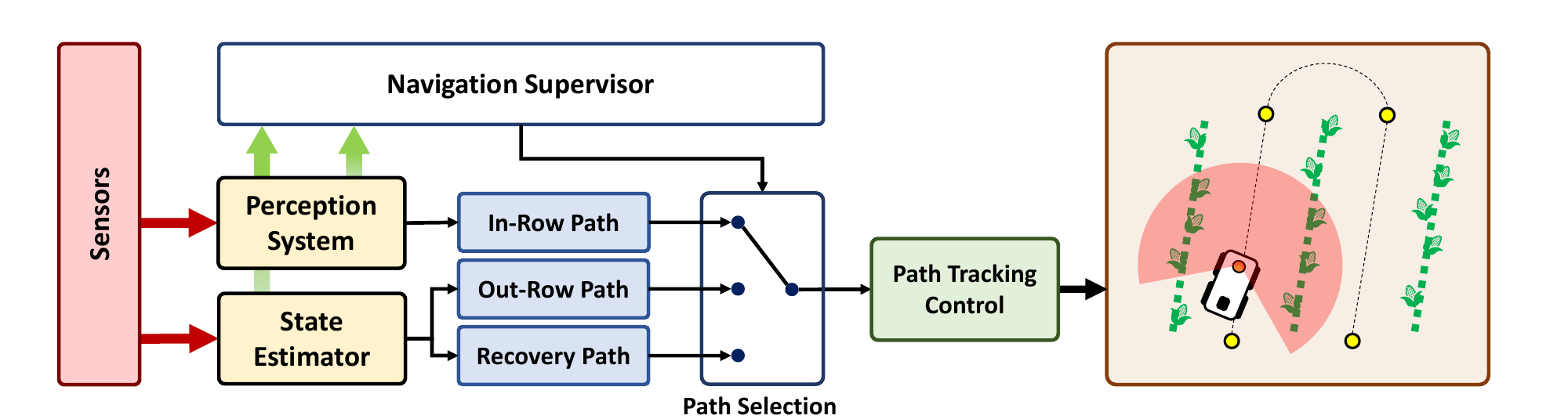}
    \caption{CropNav system diagram. Multiple sensors are used to fuse information and generate three different paths: in-row, out-row and recovery path. A navigation supervisor receives information from the perception system and the state estimation. Based on this information, the supervisor selects the appropriate path to provide safe navigation in the farm environment.}
    \label{fig:system}
    \vspace{-0.2in}
\end{figure*}

\subsection{Background Work:}
\textbf{Waypoints navigation.} In \cite{bergerman2015robot}, the authors show an autonomous navigation in orchard environments. They provide a study on localization using multiple sensors, such as GNSS, LiDAR, wheel and steering encoders to localize the robot in the field. For the navigation, they create a map using landmarks on the last plant of each row and by using this map they are able to autonomous drive the robot in the field. Since they use a big robot in orchards, it does not suffer from the GNSS reliability issue. \cite{cantelli2019small} also shows a GNSS-based navigation with pre-recorded waypoints, this time in a smaller robot. However, this paper does not address the problem of GNSS loss of reliability since robot navigates in wider rows with enough space for errors. Similarly, \cite{kayacan2018high} presents a GNSS-based navigation with moving horizon estimator and model predictive controller. They show that, assisted by RTK in ideal conditions, the system can get up to 2~cm average accuracy, which would be enough to navigate in between narrow rows. However, the problem remains unsolved for small robots that navigate under canopy.

\textbf{Row following navigation.} As demonstrated in \cite{higuti2019under}, GNSS-based navigation is not suitable for small robots in under canopy settings. Even when aided with RTK corrections, such systems cannot compensate for signal loss, accuracy loss and multi-path, surpassing the feasible error margin that a crop row can allow. In terms of under canopy navigation in physical robots, \cite{higuti2019under} and \cite{velasquez2021multi} perform extensive tests in real agricultural environments, and show that the use of LiDAR is enough to bring a safe navigation for narrow spaces. \cite{sivakumar2021learned} shows that LiDAR is not necessary for row following navigation, and a simple camera allied with a convolutional neural network can predict the distance and angle values in relation to a crop row, which can be used for under canopy navigation. This method is extensively tested in real crop, and shows robustness against illumination and some failure modes that are commonly seem in classical vision methods.

\textbf{Hybrid approaches.} In terms of multiple sensor fusion, \cite{emmi2021hybrid} presents a method that uses RGB camera and LiDAR whose fusion leads to a localization algorithm based on a topological map. Although the results show the robot can localize itself in a real agricultural environment, only offline validation is performed. In \cite{ball2016vision}, a systems that allies camera, GNSS, and inertial sensor is presented to avoid obstacles in agricultural environments and follow designated paths. Since the system is demonstrated on a real tractor, the algorithm is not suitable for under canopy scenario.



\section{MAIN CONTRIBUTION: SYSTEM DESIGN}

We present our modular system called CropNav (shown in Fig. \ref{fig:system}) to solve the problem of autonomous navigation of compact robots in under-canopy agricultural environments. The system consists of a pose estimator using a moving horizon estimator (MHE) and an Extended Kalman Filter (EKF), an under-canopy perception system based on LiDAR, a binary classifier based on LiDAR perception to check if the robot is inside or outside the crop row, a navigation failure detection system to know when the robot collides or loses traction, a high level navigation supervisor to generate the reference path for the robot based on the other modules and a unified model predictive controller (MPC) to track the reference path commanded by the supervisor. We make the following assumptions in our system to exploit the planting structure in traditional agricultural environments.

\begin{itemize}
    \item The environment is static without moving obstacles and hence the pre-recorded waypoints represent a feasible reference path that the robot can follow; 
    \item The crops are planted in parallel rows and the robot can follow the crop rows using a perception system when GNSS reliability degrades; 
    \item GNSS and perception system for row following are complementary i.e. GNSS reliability is high when in open space and during headland turn while it can be low during row following navigation.
\end{itemize}

We describe each module of our system in detail below. 

\subsection{State Estimation Algorithm}

To estimate the robot pose in real-time, we use a nonlinear MHE (\cite{kayacan2018high,gasparino2022wayfast}) followed by an EKF. The MHE uses the system model shown in Eq. \eqref{eq:kino-model}, where $x$ is the vector composed by the 2d position $p_x$ and $p_y$, and the heading angle $\theta$. The coefficients $\mu$ and $\nu$ are values in the control channel and are related to the traction dynamics. They allow us to estimate how the robot behaves when action commands are applied to the robot. When the robot moves perfectly according to the commanded actions, $\mu$ and $\nu$ should be equal to one and when the robot is stuck in place, they should be equal to zero independent of the control inputs. The system inputs $v$ and $\omega$ in Eq. \eqref{eq:kino-model} are the forward and angular velocities commanded to the wheels while \cite{kayacan2018high} used the linear and angular velocities obtained from wheel encoder measurements as input to the model in MHE. Our modified formulation enables us to detect navigation failures using the coefficients $\mu$ and $\nu$ which is not possible with the use of encoder measurements since they do not affect $\mu$ and $\nu$ when the wheels do not spin during crashes.

\vspace{-0.1in}

\begin{equation}
    \dot{x}(t)
    =
    \begin{bmatrix}
        \dot{p}_x(t) \\
        \dot{p}_y(t) \\
        \dot{\theta}(t)
    \end{bmatrix}
    =
    \begin{bmatrix}
        \mu \cdot cos(\theta(t)) & 0 \\
        \mu \cdot sin(\theta(t)) & 0  \\
        0 & \nu
    \end{bmatrix}
    \begin{bmatrix}
        v(t) \\
        \omega(t)
    \end{bmatrix}
    \label{eq:kino-model}
\end{equation}

The MHE is designed to fuse GNSS and compass measurements with control input commands to obtain state estimations. In addition, the MHE estimates three system parameters: the aforementioned $\mu$ and $\nu$ coefficients, and a $\Delta\theta$ value. This $\Delta\theta$ is the offset between the compass heading angle and the robot's true heading with respect to the world coordinate frame. We assume these three system parameters are constant in a short horizon $N \in \mathbb{N}$

\vspace{-0.1in}

\begin{align} \label{eq:mhe-optimization}
    \begin{gathered}
        \min_{x_{k:k+N}, m} ||x_k-\Tilde{x}_k||^2_{P_x} + 
        ||m-\Tilde{m}||^2_{P_m} + \\
        \sum_{i=k}^{k+N} ||y_i-h(x_i,z_i,\Delta\theta)||^2_{P_w}
    \end{gathered}
\end{align}
subject to the constraints $x_{k+1} = f(x_k,u_k)$, $\mu, \nu \in [0,1]$, and $\Delta\theta \in [-\pi,\pi)$ \cite{gasparino2022wayfast}.

Eq. \eqref{eq:mhe-optimization} shows the optimization problem in the MHE formulation, where $x$ is the state vector, $m=[\mu, \nu, \Delta\theta]^\top$ is a vector of parameters, $\Tilde{x}_k$ is the initial estimated state vector, and $\Tilde{m}$ is the estimated vector of parameters from the previous iteration. The measurement model is defined as
\begin{equation*}
    y_k-h(x_k,z_k,\Delta\theta) =
    \begin{bmatrix}
        p_{x_k} - z_{x_k} \\
        p_{y_k} - z_{y_k} \\
        \theta_k - (z_{\theta_k} + \Delta\theta)
    \end{bmatrix}
\end{equation*}
where $z_k = [z_{x_k}, z_{y_k}, z_{\theta_k}]^\top$, $z_{x_k}$ and $z_{y_k}$ are the pair of measured position coordinates from GNSS, and $z_{\theta_k}$ is the measured heading angle from the embedded inertial sensor.

The subsequent EKF uses a prediction and a measurement model to estimate the robot's states in a 2.5d space (\cite{thrun2002probabilistic, farrell2008aided}). We disregard the state associated with the height and assume a planar navigation with 3d rotation. Compared to the MHE, the EKF runs much faster. The MHE runs synchronized with the GNSS at a low frequency, while the EKF prediction model is synchronized with the embedded inertial sensor that fills the predictions at a high frequency.

\textbf{Prediction step.} In the prediction step, we use a 3d body dynamic model with constant mass such that $\hat{x}_{k+1}=f_{body}(\Tilde{x}_k,\Tilde{}{u}_k)$. The model input $\Tilde{u}=[\omega_x, \omega_y, \omega_z, a_x, a_y, a_z]^\top$ is composed by the gyroscope angular velocities $\omega_x$, $\omega_y$, $\omega_z$, and the accelerometer linear accelerations $a_x$, $a_y$, $a_z$. The predicted state vector $\hat{x} = [\hat{p}_x, \hat{p}_x, \hat{\alpha}, \hat{\beta}, \hat{\theta}]^\top$ consists of the 2d positions $\hat{p}_x$, $\hat{p}_x$, and the 3d rotation $\hat{\alpha}$, $\hat{\beta}$, $\hat{\theta}$. This vector is expressed in relation to a static world's coordinate frame.

\textbf{Correction step.} The EKF performs the correction whenever the MHE outputs are available. We augment the MHE output with the estimated pitch and roll angles $\alpha$, $\beta$ from our inertial sensor. The result is the vector $\hat{z} = [p_x, p_y, \alpha, \beta, \theta]^\top$. Note that the MHE outputs are $p_{x_{k}}$, $p_{y_{k}}$, $\theta_{k}$. The error vector used in for correction is calculated as $\hat{y} = \hat{x}-\hat{z}$. The overall process follows the well known EKF formulation \cite{thrun2002probabilistic}. The vector $\Tilde{x} = [\Tilde{p}_x, \Tilde{p}_x, \Tilde{\alpha}, \Tilde{\beta}, \Tilde{\theta}]^\top$ is the result of the correction step, which is used as the state estimation for navigation purposes. More information is available in \cite{gasparino2022wayfast}.

\subsection{Navigation Failure Detection}
 Due to its small size, under-canopy robots can face unforeseen events such as GNSS reliability changes, occlusion/anomalies in perception system and uncertainties in traction which could cause the robot to collide into plants or get stuck in place. In such cases, robot must detect such situations and try to recover from them. We use the coefficient $\mu$ estimated in our kinodynamic model by the MHE algorithm to detect collision and loss of traction. We define a threshold $\mu_{failure}$ such that if $\mu < \mu_{failure}$, then a signal is sent to trigger the recovery mode. While most of time is spent inside the crop, a region with significant failure rate is the headland, where GNSS accuracy greatly varies. Also, the terrain may have grass, water pipes, and gravel, making it a challenging area for navigation. To account for this, for all our experiments, we use $\mu_{failure} = 0.2$. This value was empirically determined and can be used as a hyper-parameter to fine tune the collision detection sensitiveness. Higher values will make the system more sensitive to collision detection.

\subsection{Under-Canopy Perception}
Row following requires its own perception system since GNSS reliability drops under plant canopy. Our perception system is based on the LiDAR-based perception algorithm described in \cite{higuti2019under,velasquez2021multi}. The main difference in our proposed system is an unified MPC controller for both navigation modes - row following and headland turn - instead of the linear controller shown in \cite{higuti2019under,velasquez2021multi} for row following. The estimated lane heading angle $\phi_{k-1}$, initialized as zero, suitably rotates the given $p_{LiDAR}^k$, the 2d point cloud at instant $k$, from the LiDAR sensor to have rows aligned with x-axis. Two rectangular bounding boxes filters out $p_{LiDAR}^k$ towards two sets, one for left side and another for right side. Such boxes are dictated by histogram peaks in y-axis and previous estimates of distance relative to the crop lane $d_{lane,k-1}$. For each set, we apply the least-squares regression to find the line that best fits them, which in turn represents respective row. Using such lines, we calculate the robot's distance to the center and angle with respect to the crop lane $\Tilde{x}_{PL} = [\Tilde{d}_{lane}, \Tilde{\phi}]^\top$.


Between two iterations of crop rows estimation algorithm, an EKF runs prediction steps using the kinodynamic model shown in Eq. \eqref{eq:kino-model}. We define this process as $\hat{x}_{PL,k} = f_{PL}(x_{PL,k-1}, u_k)$ such that $x_{PL} = [d_{lane}, \phi]^\top$. A state update $\Tilde{x}_{PL}$ from crop rows estimation algorithm happens when such estimates meet certain criteria: 1) Individual line fitting quality given by number of points and length of given set, and continuity of its angle and distance to robot; 2) Consistency with respect to known lane width. When both sides fail the validation, we skip the update step and use only the predicted values from the model. When a single side is valid, only that estimation is further used.
This enables our system to follow the lane even if there is only crop row on one side; this is not possible in monocular vision based methods such as \cite{sivakumar2021learned}. Extensive explanation and discussion about lateral distance estimation is available in \cite{higuti2019under,velasquez2021multi}.

\textbf{In-row/ Out-row classification.} A classifier is necessary to detect if the robot is inside or outside the crop rows to switch between row following and headland turning modes. A heuristic based classifier uses a threshold $N_{inrow}$ to check if the number of LiDAR points in region of interest $\hat{p}_{LiDAR}^k \geq N_{inrow}$ to decide if the robot is under canopy. $N_{inrow}$ was empirically chosen to trigger \textit{out of crop} when there is not enough points to extract the lateral lines. For corn, we determined it as 50 since it roughly relates to 0.3~m away from end of lane.

\subsection{High Level Supervisor}

The high level supervisor consists of an algorithmic block that receives the data from the state estimator and the perception algorithm to identify the current mode in which the robot is. We specify three different navigation modes: in-row, out-row, and recovery maneuver. Based on the current mode, the navigation supervisor selects the correspondent reference path which is then followed by the unified path tracking controller. The highest priority is given to the failure detection. If a failure is detected by the state estimator, the supervisor triggers the recovery mode. As a second priority, the supervisor is responsible to choose among in-row and out-row methods. The LiDAR perception systems is constantly running and checking if the robot is inside the crop rows. In the affirmative case, the supervisor sends the path that represents the middle of the row. In the negative case, the supervisor sends the path to follow the predefined path created using recorded waypoints.

\subsection{Reference Generation}

\begin{figure}[htp]
    \centering
    \includegraphics[width=0.8\linewidth]{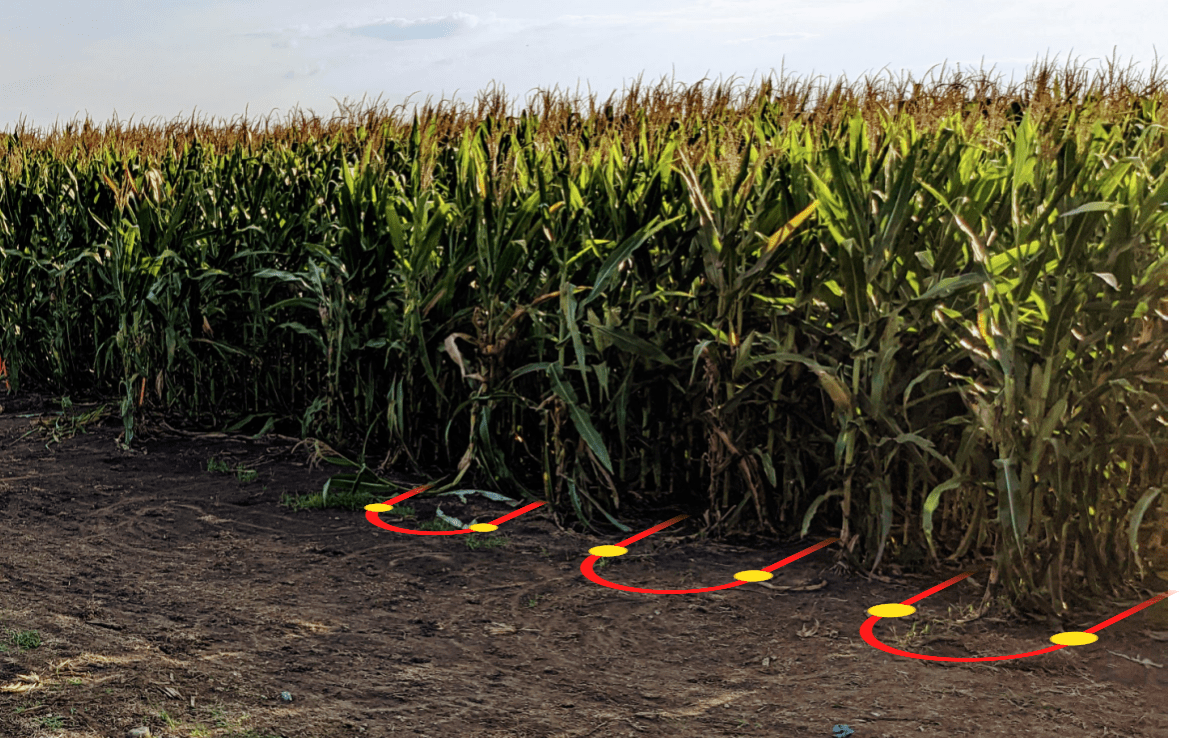}
    \caption{\textbf{Waypoints recording}. Yellow circles are examples of how the waypoints are recorded. The red lines represent the reference path automatically created from these points.}
    \label{fig:waypoints-example}
    \vspace{-0.1in}
\end{figure}

The three options of reference state trajectories created during navigation are: GNSS-based path, perception-based path, and recovery path. The \textbf{GNSS-based path} is generated by using pre-recorded waypoints. The waypoints may be created by either driving the robot manually to the desired points, or can be accessed from a crop map. Such map, frequently with RTK-GNSS accuracy, is usually created by the tractor that plants the crop. For each navigated crop row, the system requires one waypoint per entry and one per exit. Fig. \ref{fig:waypoints-example} shows an example of how to define the waypoints.

The \textbf{perception-based path} is created to follow the crop rows. It is defined as a straight line that represents the middle of the row transformed to the robot's coordinate. The \textbf{recovery maneuver path} is a buffer that contains the path states estimated using the EKF algorithm. This buffer accumulates the previous seconds of estimated states, so in case of a failure, the robot can try to recover by redoing a path that we know is safe (since it was already followed). In agriculture, most of the failures happen because of instantaneous obstructions of sensors or measurement noise. By redoing this same path the robot can successfully recover from collisions most times as we show in our experiments.

\subsection{Path Tracking Controller}

\begin{figure}[htp]
    \centering
    \includegraphics[trim={6cm 2cm 6cm 2cm}, clip, width=1.0\linewidth]{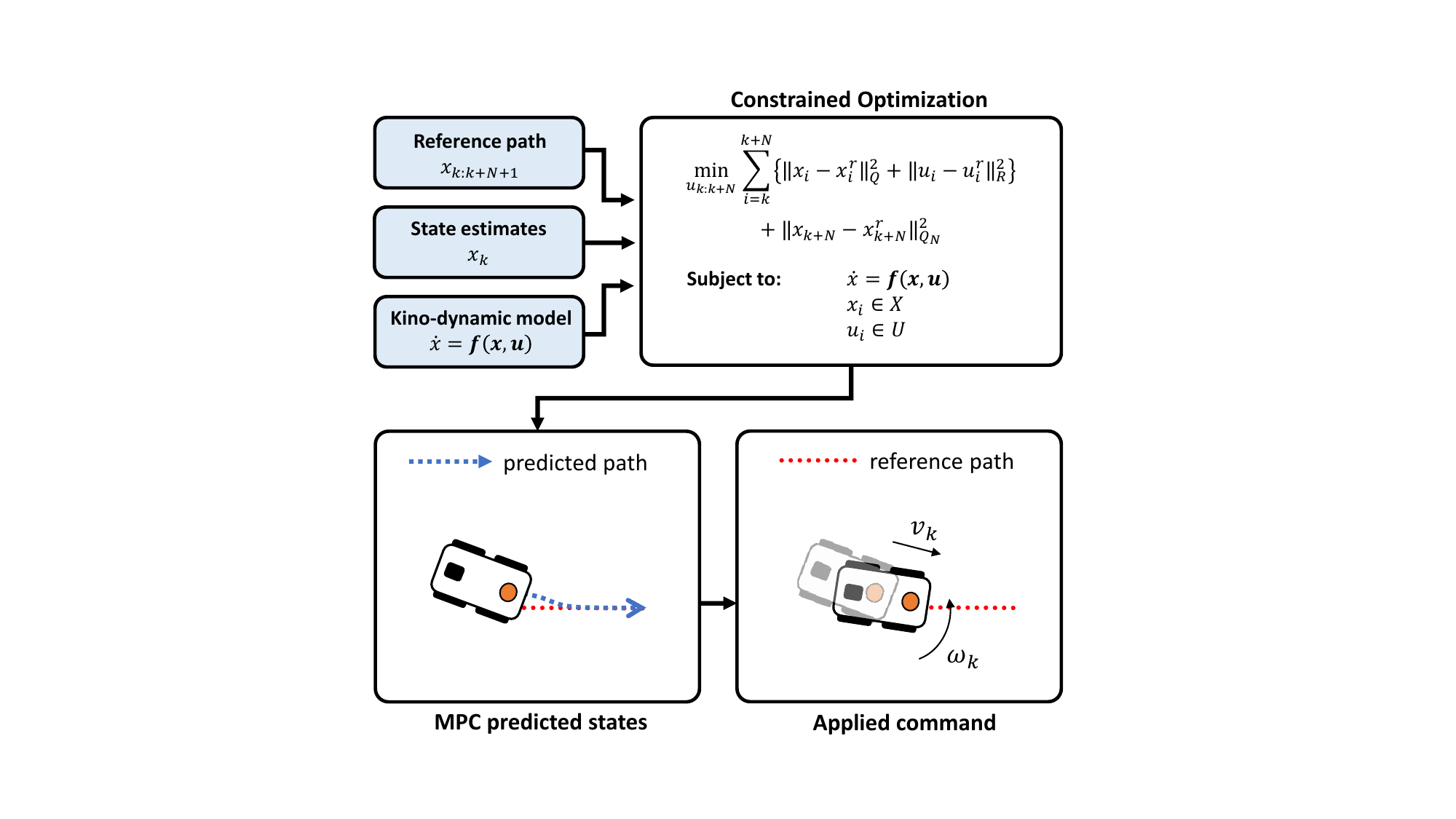}
    \caption{\textbf{Model predictive control}. The diagram explains the controller responsible to track a reference state path.}
    \label{fig:mpc-control}
    \vspace{-0.2in}
\end{figure}

Running parallel to the state estimator, we present a unified MPC that drives the robot to track a given reference path, as shown in Fig. \ref{fig:mpc-control}. We take advantage of the MPC capabilities to satisfy states and input constraints to find a solution to a minimization problem. Similar to our MHE formulation, the MPC uses the kinodynamic model described in Equation \eqref{eq:kino-model}. The states $p_{x}$ and $p_{y}$ represent the robot's location in a 2d space, $\theta$ is the heading angle, $\nu$ and $\mu$ are the angular and linear coefficient parameters respectively, and $\omega_k$ and $v_k$ are the angular and linear velocities respectively.

Because the robot is a four motor drive skid-steer robot, we constrain the maximum allowable speed for each individual motor. Motors at the same side are subject to the same speed commands; $v_{left}$ is the speed for left side, and $v_{right}$ for right side. We calculate $v_{left}$ and $v_{right}$ in terms of the MPC control outputs as $v_{left} = v_k / \mu - L \omega_k / 2$ and $v_{right} = v_k / \mu + L \omega_k / 2$ where $L$ is the track width of the robot. We define the control set $\mathbb{U}$ such that $\minus v_{max} \leq v_{left} \leq v_{max}$ and $\minus v_{max} \leq v_{right} \leq v_{max}$, where $v_{max}$ is the maximum allowable wheel speed. The states set $\mathbb{X}$ is not constrained and represents the entire 3d set of real numbers $\mathbb{R}^3$.

To design the online reference tracking MPC, we choose an optimization horizon $N \in \mathbb{N}$ and positive definite matrices $Q$ and $R$. The following finite horizon optimization formulation is solved to obtain the current control action
\begin{align} \label{eq:mpc-optimization}
    \begin{gathered}
        \min_{u_{k:k+N \minus 1}} \sum_{i=k}^{k+N \minus 1} \left\{ ||x_i \minus x_i^r||^2_Q + ||u_i \minus u_i^r||^2_R \right\} \\
        + ||x_{k+N} \minus x_{k+N}^r||^2_{Q_N}
    \end{gathered}
\end{align}
such that, at every iteration, the optimization framework is subject to the constraints $x_k \in \mathbb{X}$ and $u_k \in \mathbb{U}$. As a result of the optimization problem \eqref{eq:mpc-optimization}, we obtain the control sequence $u_{k:k+N \minus 1}$. We take the first element $u_{k}$ and apply to the motors to make the robot follow the reference path.

\section{EXPERIMENTAL RESULTS}

We performed extensive experiments in a real agricultural field with our presented system deployed to a physical robot. They are divided into four types: serpentine path (580~m) where robot navigates between crop rows and switches to next lane, same path without recovery mechanism, same path using only GNSS, and a longer path (1.2~km) with similar pattern. In such experiments, we recorded the human interventions and recoveries that happened along these paths.

\begin{figure}[htp]
    \centering
    \includegraphics[width=\linewidth]{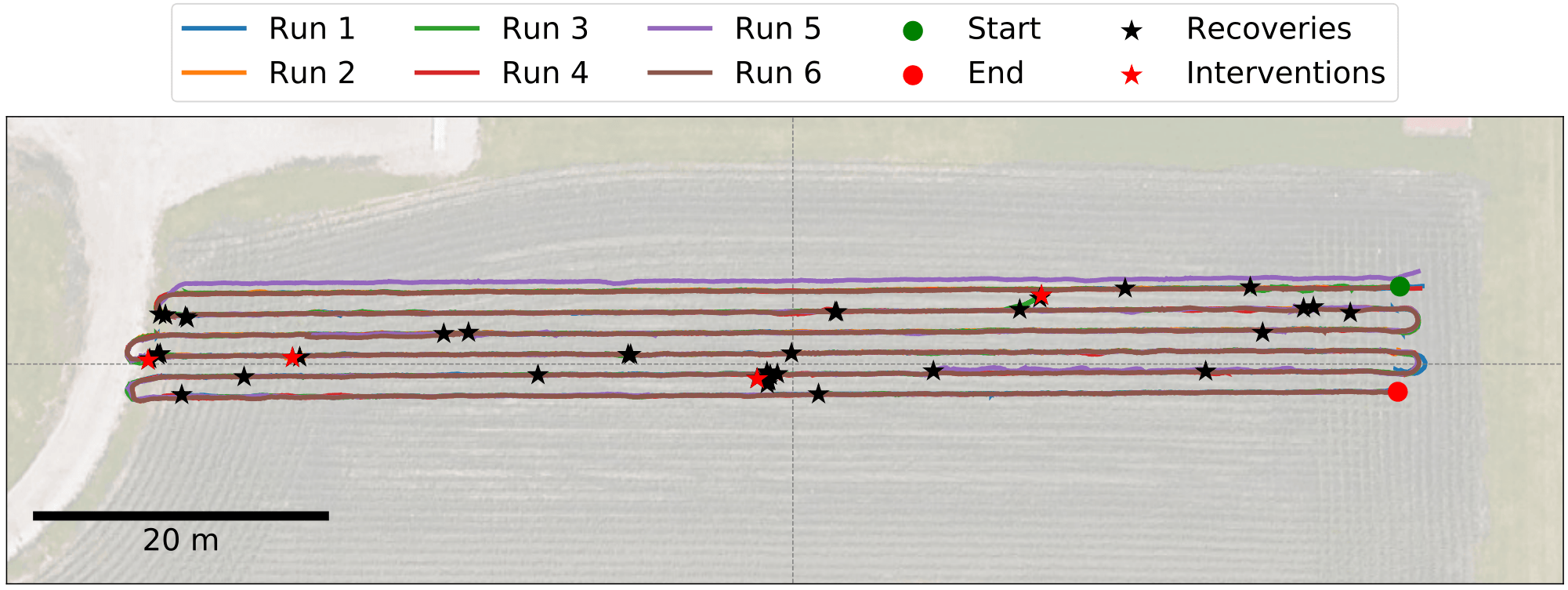}
    \caption{\textbf{Experiment with CropNav.} In this experiment, our method runs trough six crop rows. Plot shows the navigated path with recoveries and failures locations.}
    \label{fig:result-ours}
    \vspace{-0.1in}
\end{figure}

\textbf{Presented Method.} The first set of experiments presents our method. In this experiment, we recorded a sequence of 12 waypoints, each specifying the entrance or termination of a crop row. The robot started close to the first waypoint and autonomously navigated to complete the entire sequence of points. We saved the state estimates, sensors readings, recovery and interventions locations. We repeated this same experiment six times. Fig. \ref{fig:result-ours} shows a plot containing these six repetitions with an overall distribution of recoveries and intervention points. For each repetition, the robot navigated about 580~m. In the total, the system needed four human interventions, an average of 885~m per intervention. Note this is higher than any row-following navigation alone can achieve, since the row length is about 90~m. In addition, we can see the recoveries did not happen concentrated in a single location, which shows that different sources of errors can affect the navigation system.

\begin{figure}[htp]
    \centering
    \includegraphics[width=\linewidth]{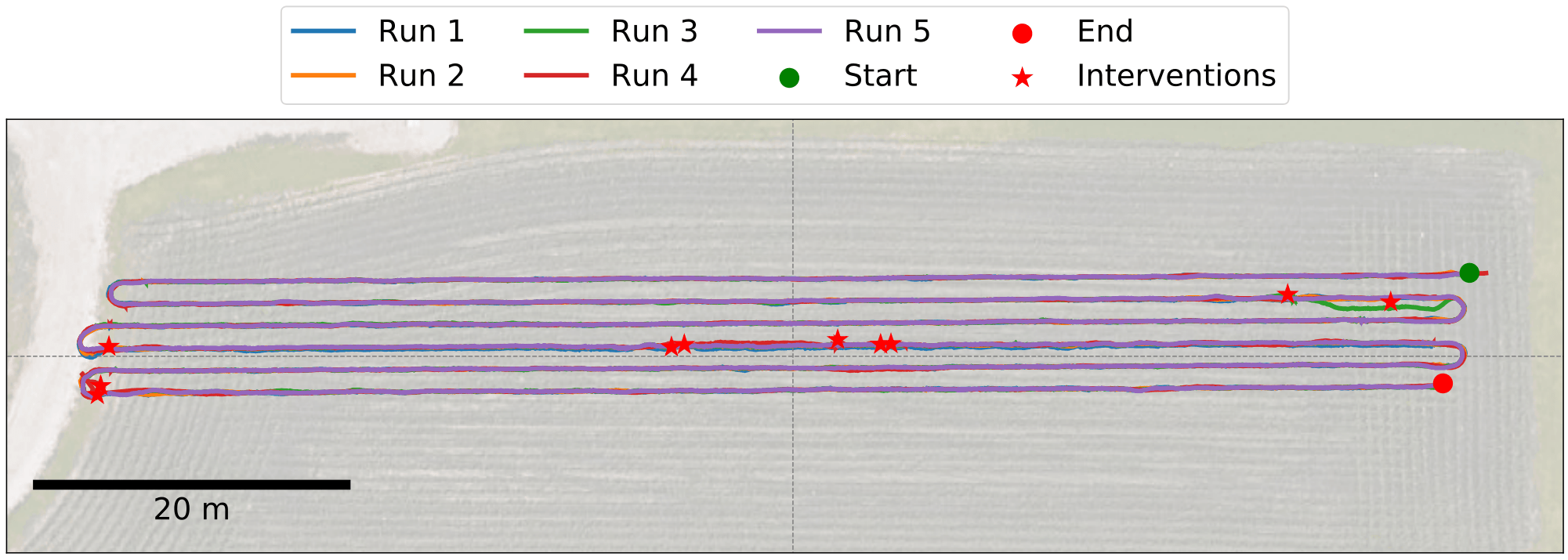}
    \caption{\textbf{Navigation without recovery.} We verify how the robot behaves when relying only on the in-row and out-row navigation. The trajectory length is approximately 580~m.}
    \label{fig:result-wo-recovery}
    \vspace{-0.1in}
\end{figure}

To verify the importance of the recovery algorithm, we performed experiments where this mechanism was disabled, thus the navigation relied solely on the in-row and out-row navigation methods. As shown in Fig. \ref{fig:result-wo-recovery}, we tried five runs using the same recorded waypoints as shown in the previous experiment. As we can see, the overall number of human interventions drastically increased, requiring a person to always recover the robot after failures. The metric of 885~m per intervention decreased to 290~m per intervention during this experiment.

\textbf{GNSS Only Navigation.} As a baseline, we used a GNSS-only navigation coupled with our recovery algorithm. The intuition for this experiment is to show that the row following navigation is necessary, since GNSS loss of reliability causes robot failures. We show through these experiments that the robot was able to navigation for two entire rows, with collisions that recovered well using our recovery algorithm. As the robot entered deeper into the crop, the GNSS reliability lowered, and failures became more often.

\begin{figure}[htp]
    \centering
    \includegraphics[width=\linewidth]{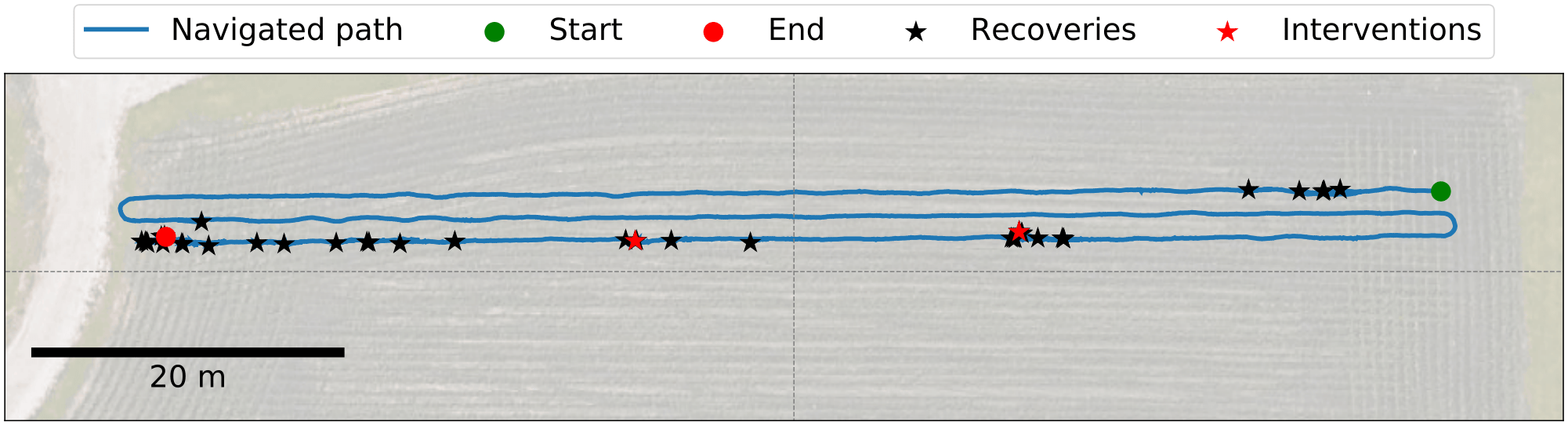}
    \caption{\textbf{Navigation using GNSS-based algorithm.} GNSS-based navigation is impractical. Due to the low position reliability when under canopy, the navigation constantly collides into one of the sides, triggering the recovery maneuver.}
    \label{fig:result-gnss-only}
    \vspace{-0.1in}
\end{figure}

\begin{figure}[htp]
    \centering
    \begin{subfigure}[h]{0.31\linewidth}
        \centering
        \includegraphics[width=\textwidth]{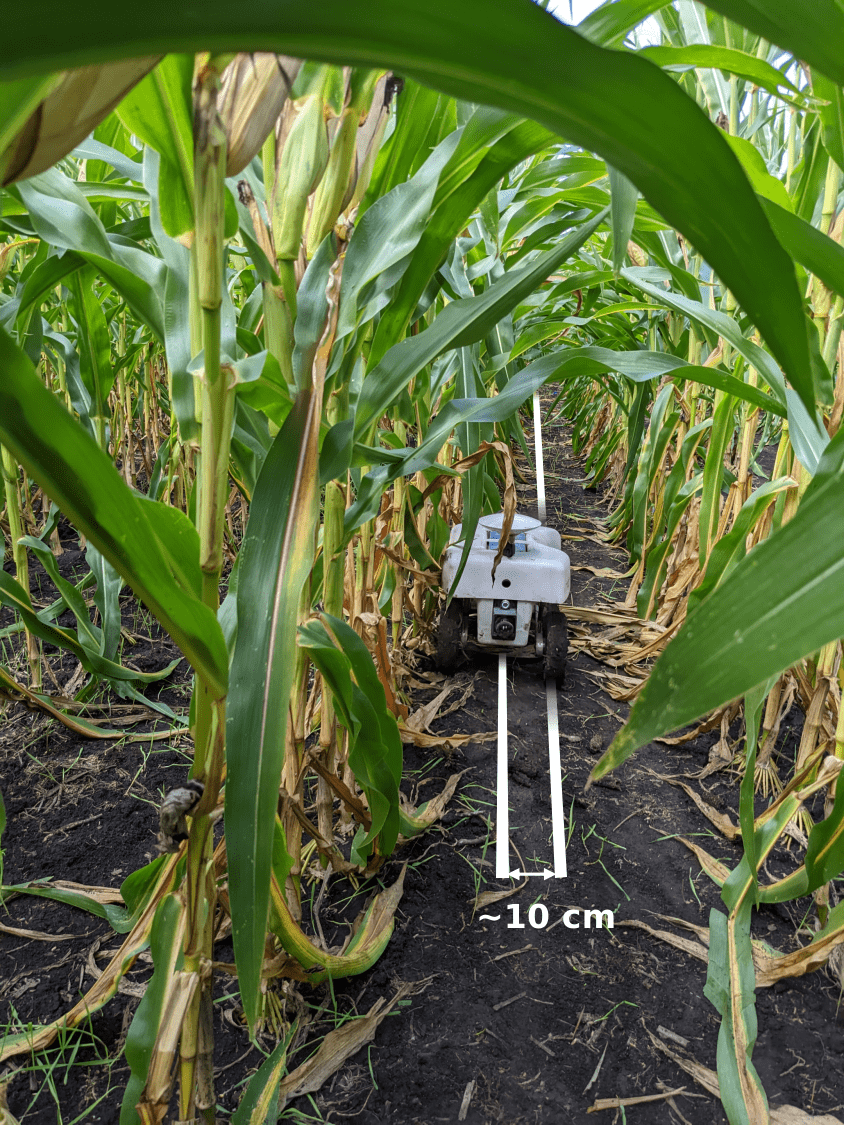}
    \end{subfigure}%
    \hfill
    \begin{subfigure}[h]{0.31\linewidth}
        \centering
        \includegraphics[width=\textwidth]{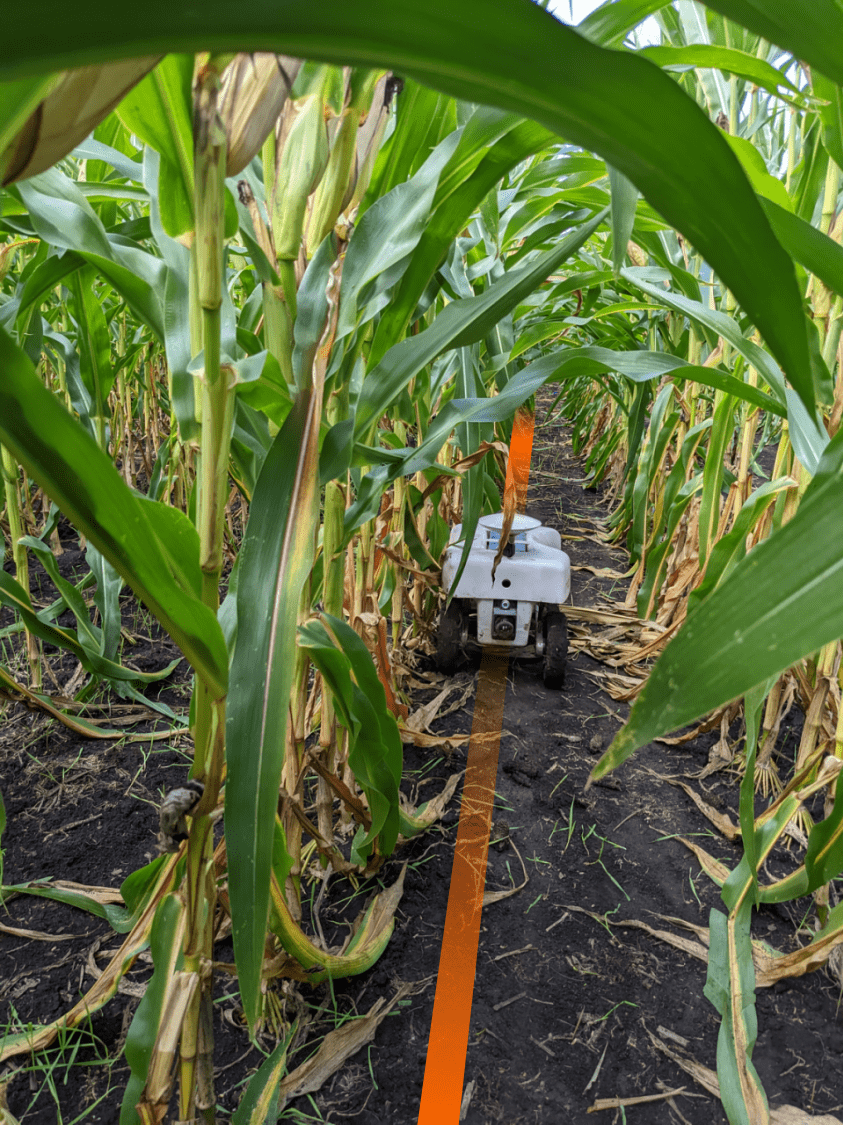}
    \end{subfigure}
    \hfill
    \begin{subfigure}[h]{0.31\linewidth}
        \centering
        \includegraphics[width=\textwidth]{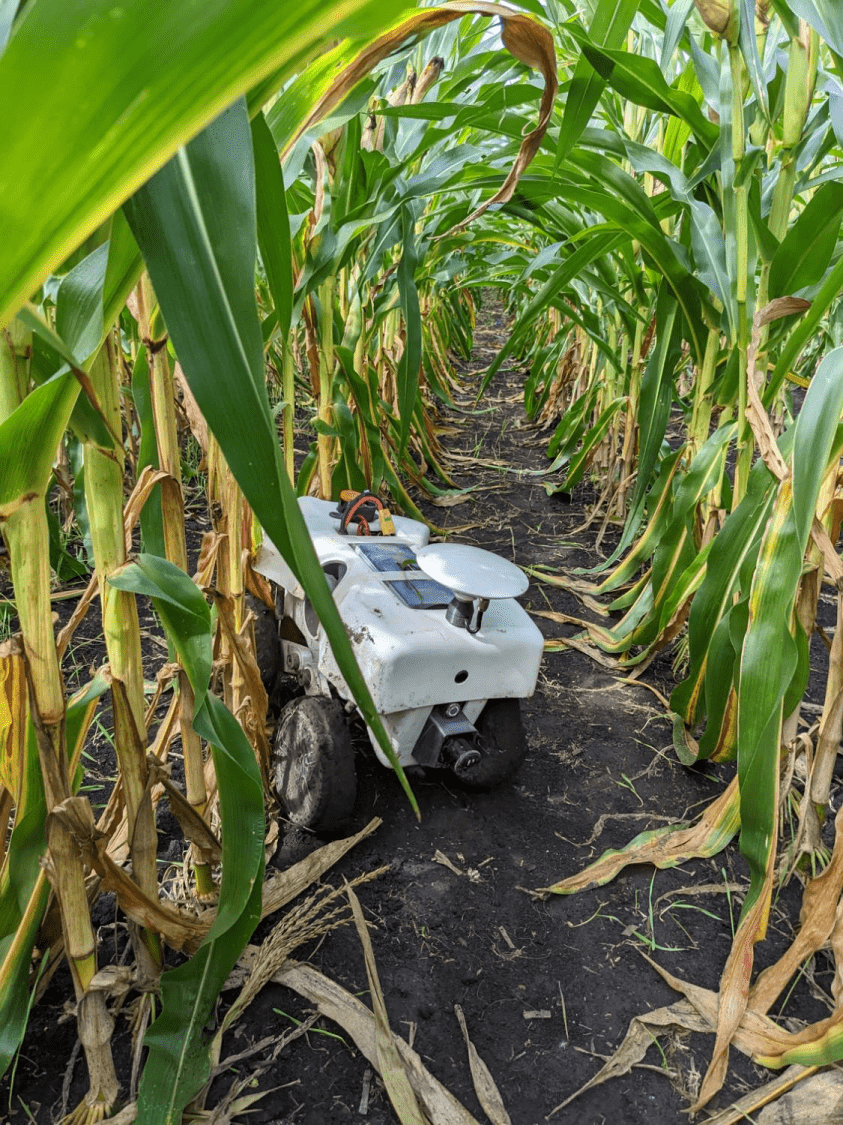}
    \end{subfigure}
    \caption{\textbf{Navigation offset generated by low GNSS reliability.} Due to multi-path, even RTK severely suffers under canopy, causing (from left to right) A) The robot navigates with a constant offset; B) The reference path shifts to the left; C) A failure mode that was frequently observed in the GNSS-only navigation due to the shifted reference path.}
    \label{fig:gps-only-experiment}
    \vspace{-0.2in}
\end{figure}

Fig. \ref{fig:result-gnss-only} shows the GNSS-only navigation experiment. In this experiment, the robot used the same waypoints recorded from the previous experiments, however, due to the high amount of failures, we decided to stop the experiment earlier to avoid damaging the crop. As we can see in Fig. \ref{fig:gps-only-experiment}, the robot was constantly driving with an offset, near the crop row to the left. Because of this, many recoveries and interventions happened in a short period of time, which shows the GNSS-only navigation is impractical for such scenarios.

\begin{figure}[htp]
    \centering
    \includegraphics[width=0.98\linewidth]{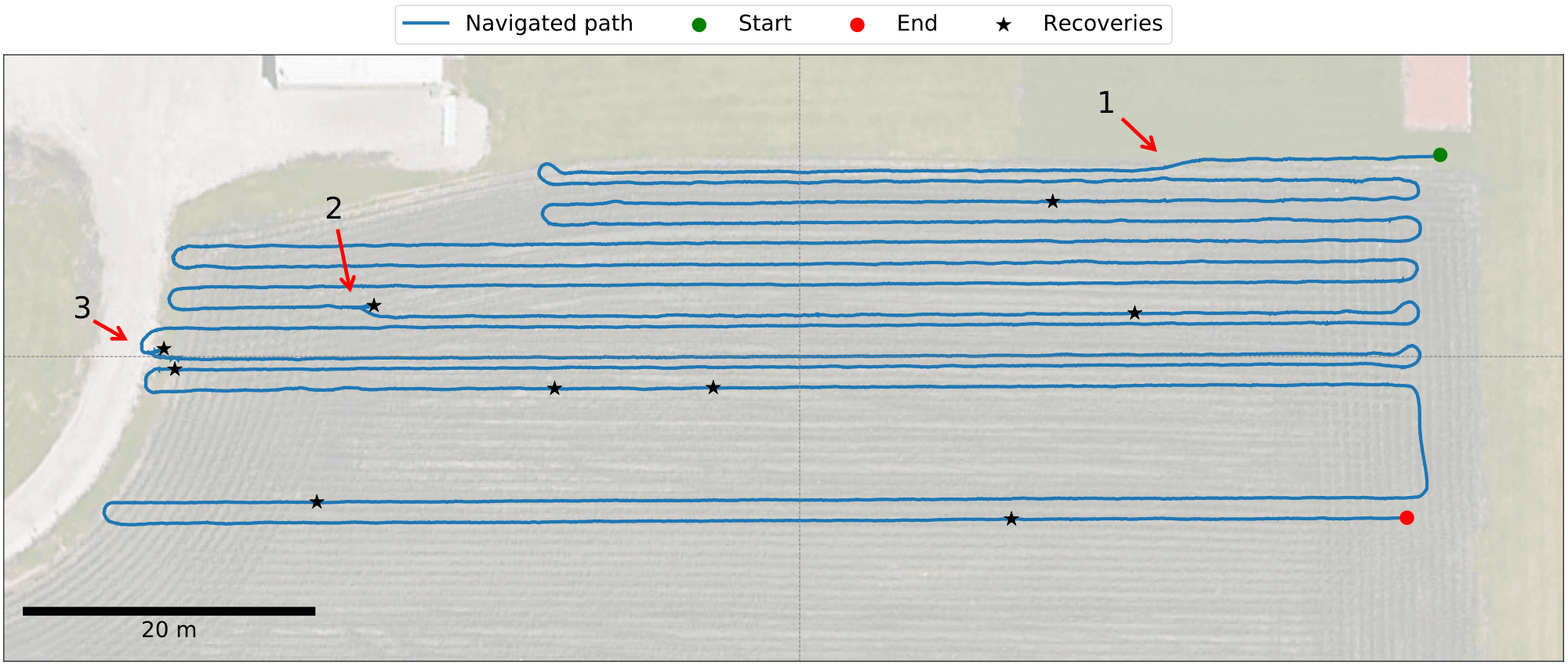}
    \caption{\textbf{Navigation in a long path.} The total navigation length is about 1.2~km without any human intervention. We highlight three events that happened during this experiments. Numbers 1-3 mark three situations detailed in the text.}
    \label{fig:long-path}
    \vspace{-0.1in}
\end{figure}

\textbf{Long Path Navigation.} In this experiment, we tested the capabilities of our navigation system in a long path. The total path length is approximately 1.2~km. We defined a sequence of 28 waypoints, each positioned in the entrance or exit of a row. Again, the robot started at the first waypoints and autonomously navigated towards the last waypoint. The path was chosen to make the robot navigate through 12 rows, leave the first group of rows to navigate using the GNSS-based navigation, and reach the last two rows. We repeated this long navigation twice, each lasting about 40 minutes. Figure \ref{fig:long-path} shows the plot of one of these tries.

We highlighted three contrasting situations that can be observed in Fig. \ref{fig:long-path}, denoted using numbers. In the first situation, due to a long gap of missing corn plants in the left corn row, the robot detected the adjacent row, which caused the navigation system to deviate the trajectory to the nearest row to the left. As we can see, no collision was caused, and the robot continued navigating as usual. In the second situation, a similar event happened, but this time it was due to a recovery maneuver. Due to a failure in the perception system, the robot recovered and used a gap to switch to the adjacent row. And finally, the third occasion was caused by a failure in the control system. As we can see in the map, there is an intersection between ground and grass at that point, so the abrupt change in the terrain properties caused an overshoot in that path. This deviation caused the robot to collide when trying to enter the row, which lead to erroneously taking the adjacent row on its right side. Once again, the recovery mechanism was able to correct this situation and the robot continued navigating without human interventions. We summarize all the experiments in Table \ref{tab:experiments-summary}.

\begin{table}[htp]
    \centering
    \resizebox{0.6\columnwidth}{!}{%
    \begin{tabular}{cccc}
        \hline
        \multicolumn{4}{|c|}{\textbf{GNSS-only navigation (w/ recovery)}} \\ \hline
        \multicolumn{1}{|c|}{\textbf{Run \#}} & \multicolumn{1}{c|}{\textbf{Recoveries}} & \multicolumn{1}{c|}{\textbf{Interventions}} & \multicolumn{1}{c|}{\textbf{Avg. {[}m / interv.{]}}} \\ \hline
        \multicolumn{1}{|c|}{\textbf{1}} & \multicolumn{1}{c|}{17} & \multicolumn{1}{c|}{2} & \multicolumn{1}{c|}{120} \\ \hline
        \rowcolor[HTML]{C0C0C0} 
        \multicolumn{1}{|c|}{\cellcolor[HTML]{C0C0C0}\textbf{Total}} & \multicolumn{1}{c|}{\cellcolor[HTML]{C0C0C0}17} & \multicolumn{1}{c|}{\cellcolor[HTML]{C0C0C0}2} & \multicolumn{1}{c|}{\cellcolor[HTML]{C0C0C0}120} \\ \hline
        \multicolumn{1}{l}{} & \multicolumn{1}{l}{} & \multicolumn{1}{l}{} & \multicolumn{1}{l}{} \\ \hline
        \multicolumn{4}{|c|}{\textbf{CropNav (ours w/o recovery)}} \\ \hline
        \multicolumn{1}{|c|}{\textbf{Run \#}} & \multicolumn{1}{c|}{\textbf{Recoveries}} & \multicolumn{1}{c|}{\textbf{Interventions}} & \multicolumn{1}{c|}{\textbf{Avg. {[}m / interv.{]}}} \\ \hline
        \multicolumn{1}{|c|}{\textbf{1}} & \multicolumn{1}{c|}{-} & \multicolumn{1}{c|}{0} & \multicolumn{1}{c|}{-} \\ \hline
        \multicolumn{1}{|c|}{\textbf{2}} & \multicolumn{1}{c|}{-} & \multicolumn{1}{c|}{2} & \multicolumn{1}{c|}{290} \\ \hline
        \multicolumn{1}{|c|}{\textbf{3}} & \multicolumn{1}{c|}{-} & \multicolumn{1}{c|}{1} & \multicolumn{1}{c|}{580} \\ \hline
        \multicolumn{1}{|c|}{\textbf{4}} & \multicolumn{1}{c|}{-} & \multicolumn{1}{c|}{6} & \multicolumn{1}{c|}{96.7} \\ \hline
        \multicolumn{1}{|c|}{\textbf{5}} & \multicolumn{1}{c|}{-} & \multicolumn{1}{c|}{1} & \multicolumn{1}{c|}{580} \\ \hline
        \rowcolor[HTML]{C0C0C0} 
        \multicolumn{1}{|c|}{\cellcolor[HTML]{C0C0C0}\textbf{Total}} & \multicolumn{1}{c|}{\cellcolor[HTML]{C0C0C0}-} & \multicolumn{1}{c|}{\cellcolor[HTML]{C0C0C0}10} & \multicolumn{1}{c|}{\cellcolor[HTML]{C0C0C0}290} \\ \hline
        \multicolumn{1}{l}{} & \multicolumn{1}{l}{} & \multicolumn{1}{l}{} & \multicolumn{1}{l}{} \\ \hline
        \multicolumn{4}{|c|}{\textbf{CropNav (ours w/ recovery)}} \\ \hline
        \multicolumn{1}{|c|}{\textbf{Run \#}} & \multicolumn{1}{c|}{\textbf{Recoveries}} & \multicolumn{1}{c|}{\textbf{Interventions}} & \multicolumn{1}{c|}{\textbf{Avg. {[}m / interv.{]}}} \\ \hline
        \multicolumn{1}{|c|}{\textbf{1}} & \multicolumn{1}{c|}{1} & \multicolumn{1}{c|}{0} & \multicolumn{1}{c|}{-} \\ \hline
        \multicolumn{1}{|c|}{\textbf{2}} & \multicolumn{1}{c|}{0} & \multicolumn{1}{c|}{2} & \multicolumn{1}{c|}{145} \\ \hline
        \multicolumn{1}{|c|}{\textbf{3}} & \multicolumn{1}{c|}{5} & \multicolumn{1}{c|}{1} & \multicolumn{1}{c|}{590} \\ \hline
        \multicolumn{1}{|c|}{\textbf{4}} & \multicolumn{1}{c|}{1} & \multicolumn{1}{c|}{1} & \multicolumn{1}{c|}{590} \\ \hline
        \multicolumn{1}{|c|}{\textbf{5}} & \multicolumn{1}{c|}{11} & \multicolumn{1}{c|}{0} & \multicolumn{1}{c|}{-} \\ \hline
        \multicolumn{1}{|c|}{\textbf{6}} & \multicolumn{1}{c|}{0} & \multicolumn{1}{c|}{0} & \multicolumn{1}{c|}{-} \\ \hline
        \rowcolor[HTML]{C0C0C0} 
        \multicolumn{1}{|c|}{\cellcolor[HTML]{C0C0C0}\textbf{Total}} & \multicolumn{1}{c|}{\cellcolor[HTML]{C0C0C0}17} & \multicolumn{1}{c|}{\cellcolor[HTML]{C0C0C0}4} & \multicolumn{1}{c|}{\cellcolor[HTML]{C0C0C0}885} \\ \hline
        \multicolumn{1}{l}{} & \multicolumn{1}{l}{} & \multicolumn{1}{l}{} & \multicolumn{1}{l}{} \\ \hline
        \multicolumn{4}{|c|}{\textbf{Long path navigation}} \\ \hline
        \multicolumn{1}{|c|}{\textbf{Run \#}} & \multicolumn{1}{c|}{\textbf{Recoveries}} & \multicolumn{1}{c|}{\textbf{Interventions}} & \multicolumn{1}{c|}{\textbf{Avg. {[}m / interv.{]}}} \\ \hline
        \multicolumn{1}{|c|}{\textbf{1}} & \multicolumn{1}{c|}{7} & \multicolumn{1}{c|}{1} & \multicolumn{1}{c|}{1200} \\ \hline
        \multicolumn{1}{|c|}{\textbf{2}} & \multicolumn{1}{c|}{9} & \multicolumn{1}{c|}{0} & \multicolumn{1}{c|}{-} \\ \hline
        \rowcolor[HTML]{C0C0C0} 
        \multicolumn{1}{|c|}{\cellcolor[HTML]{C0C0C0}\textbf{Total}} & \multicolumn{1}{c|}{\cellcolor[HTML]{C0C0C0}16} & \multicolumn{1}{c|}{\cellcolor[HTML]{C0C0C0}1} & \multicolumn{1}{c|}{\cellcolor[HTML]{C0C0C0}2400} \\ \hline
    \end{tabular}}
    \caption{\textbf{Experiments summary.} The table is divided in four experiments. The first column shows the runs per experiment, the second shows the count of recoveries performed, the third counts the number of human interventions, and the last, the average navigated distance per intervention. }
    \label{tab:experiments-summary}
    \vspace{-0.1in}
\end{table}

\section{CONCLUSIONS}

We presented an autonomous navigation system for real agricultural environments. Our approach uses a modular architecture that combines a state estimator, a model predictive controller, and a high level navigation supervisor to provide safe navigation in farm scenarios. We extensively validated our method in long paths that alternates between in and out crop situations and we demonstrated that our platform was able to accurately detect the situation and appropriately switch to the proper navigation mode. In addition, we performed experiments to show the efficacy of the fail detection system. By using a model parameter coefficient estimated in real time by our MHE, we can accurately detect when the robot collides and promote recovery maneuvers. These maneuvers can extend the autonomous navigation time and alleviate the need of human interventions. Such advancement is an important step for reliable crop inspection, high-throughput phenotyping and data collection, cover crop planting, herbicides application, and among many other applications. For future works, we believe that further improving the system to detect row entrances can dismiss the need of RTK-GNSS reliability.

\addtolength{\textheight}{-6cm}   
                                  
\bibliographystyle{IEEEtran}
\bibliography{references}

\end{document}